\documentclass{article}
\usepackage[utf8]{inputenc}
\usepackage{url}
\usepackage{graphicx,algorithm,algcompatible,amsmath, relsize}
\usepackage{hyperref}
\usepackage{subfigure}
\usepackage{algpseudocode}
\usepackage{amsthm,amssymb,inputenc,titling}
\usepackage{hyperref}
\usepackage[T1]{fontenc}

\usepackage{listings}
\usepackage[T1]{fontenc}

\usepackage{blindtext}
\usepackage{xcolor}
\definecolor{codegreen}{rgb}{0,0.6,0}
\definecolor{codegray}{rgb}{0.5,0.5,0.5}
\definecolor{codepurple}{rgb}{0.58,0,0.82}
\definecolor{backcolour}{rgb}{0.95,0.95,0.92}

\lstdefinestyle{mystyle}{
    backgroundcolor=\color{backcolour},
    commentstyle=\color{codegreen},
    keywordstyle=\color{magenta},
    numberstyle=\tiny\color{codegray},
    stringstyle=\color{codepurple},
    basicstyle=\ttfamily\footnotesize,
    breakatwhitespace=false,
    breaklines=true,
    captionpos=b,
    keepspaces=true,
    numbers=left,
    numbersep=5pt,
    showspaces=false,
    showstringspaces=false,
    showtabs=false,
    tabsize=2
}
\usepackage{blindtext}
\title{Multi-Task Learning of Query Intent and Named Entities using Transfer Learning}
\author{
Shalin Shah
\texttt{\{shalin.shah@target.com\}}\\
Ryan Siskind
\texttt{\{ryan.siskind@target.com\}}\\
\\
AI Sciences\\
Target Corporation\\
Sunnyvale, CA 94086, USA \\
}
\date{}
\begin{document}
\maketitle
\begin{abstract}
Named entity recognition (NER) has been studied extensively and the earlier algorithms were based on sequence labeling like Hidden Markov Models (HMM) and conditional random fields (CRF). These were followed by neural network based deep learning models. Recently, BERT has shown new state of the art accuracy in sequence labeling tasks like NER. In this short article, we study various approaches to task specific NER. Task specific NER has two components - identifying the intent of a piece of text (like search queries), and then labeling the query with task specific named entities. For example, we consider the task of labeling Target store locations in a search query (which could be entered in a search box or spoken in a device like Alexa or Google Home). Store locations are highly ambiguous and sometimes it is difficult to differentiate between say a location and a non-location. For example, "pickup my order at orange store" has "orange" as the store location, while "buy orange at target" has "orange" as a fruit. We explore this difficulty by doing multi-task learning which we call global to local transfer of information. We jointly learn the query intent (i.e. store lookup) and the named entities by using multiple loss functions in our BERT based model and find interesting results.
\end{abstract}
\section{Introduction}
We use BERT \cite{devlin2018bert} for multi-task learning. We jointly learn the intent of the search query (store lookup) and also the named entities (store locations). We find that this method performs better in terms of precision (which is more important than recall). The false positive rate improved significantly through the global query intent loss function combined with a tagging loss function. As far as we know, this is the first attempt to jointly learn task specific NER along with a text classification loss function. We found one other piece of work that does something similar using LSTM networks \cite{ma2017jointly}. A good overview of multi-task learning can be found in \cite{ruder2017overview} and \cite{zhang2017survey}. \href{https://medium.com/walmartglobaltech/joint-intent-classification-and-entity-recognition-for-conversational-commerce-35bf69195176}{This page} does something very similar using BERT but its not completely clear on whether their work uses multi-task learning.\\\\
Also, a distinguishing feature of our work is that each word has its own intent, rather than just one binary output for the intent (which appears similar to the ideas in \cite{vaswani2017attention}). Several intents can be learnt together with the respective named entities, like navigational store hours lookup (store name is the entity), product lookup (the product is the entity), add to cart (product is the entity), store pickup (the store is the entity), page navigation (the page location is the entity, like navigating to a category page etc.), product styles (in which the style and the product would be the entities). All intents have their own token level loss functions for intent along with a common named entity recognition engine in a multi-task setting (\href{https://ai.googleblog.com/2019/10/exploring-massively-multilingual.html}{This page} has several interesting ideas about multi-task learning. In particular they use the term negative transfer in which one loss function cannibalizes the effect of the other and both then do poorly).\\\\
BERT \cite{devlin2018bert} is a very effective pre-trained model which can be fine tuned for various tasks. BERT learns a masked language model in which a few words of a piece of text are masked and the model learns how to predict them. It is bidirectional and uses attention \cite{vaswani2017attention}, which includes interactions of a word with other words following it or preceding it. By adding task specific loss functions and learning on a comparatively smaller set of data, the model can be fine tuned. For example, the final layer of the model could be a Softmax layer which can do text classification or tagging. Or it could possibly be a linear layer.\\\\
We use BERT with two added loss functions - one that learns NER tags and the other that learns a classification problem. The NER task learns store locations and the classification task learns the query intent (i.e. store lookup). More in section 4.
\section{Named Entity Recognition and Sequence Labeling}
Sequence labeling is a task in which a piece of text has to be tagged with information like names, parts of speech tags, and other tags. Named entity recognition could be a task that follows POS tagging and NP chunking, but it does not have to be.\\\\
The earliest NER models were based on sequence labeling, like hidden Markov models and conditional random fields. The NLTK package \cite{loper2002nltk} and the Stanford NLP package \cite{manning2014stanford} both use CRF for sequence tagging. SpaCy \cite{spacy2} \cite{srinivasa2018natural} uses convolutional neural networks and recurrent neural networks to do named entity recognition.\\\\
Recently, BERT has shown to obtain state of the art (SOTA) results on named entity tasks. The HuggingFace library \cite{wolf2019huggingface} provides a transformers python package which includes pre-trained models for various language tasks like named entity recognition. We use this package in our work along with BERT \cite{devlin2018bert} and pyTorch \cite{paszke2019pytorch}.\\\\
\href{https://github.com/sebastianruder/NLP-progress/blob/master/english/named_entity_recognition.md}{This git page} does a comprehensive survey of SOTA algorithms for various NLP tasks including named entity recognition.
\section{Multi-task Learning}
Multi-task learning has multiple objective functions which makes it a multi-objective learning problem. The most common approach to multi-objective learning is to just do a convex combination of the multiple objectives with weights for each objective while sharing the neural network representation for each objective. This has shown to improve generalization as described in this book \cite{caruana1997multitask}.\\\\
It is natural that learning the intent of a query and then doing NER will result in a hierarchical model which will reduce the false positive rate. Learning the query intent jointly with NER will change the representation (neural network) for NER tagging as well and might result in a better precision.\\\\
In a multi-task or a multi-objective problem, the inductive bias of the classifier is shared between multiple objectives \cite{caruana1997multitask}. This results in an amortized classifier, which, for multiple-objectives, only differs in the final layers. This results in an improved performance because of the global to local transfer.\\\\
In our case, the global information for a query (the query intent), and the word specific local information (the NER labels) are jointly learnt, while sharing the entire BERT neural network model.
\section{The NER Algorithm}
As figures 1 and 2 show, there are two distinct loss functions which share a common neural network.
\begin{figure}[ht]
\label{ner-examples}
\centering
\includegraphics[scale=0.6]{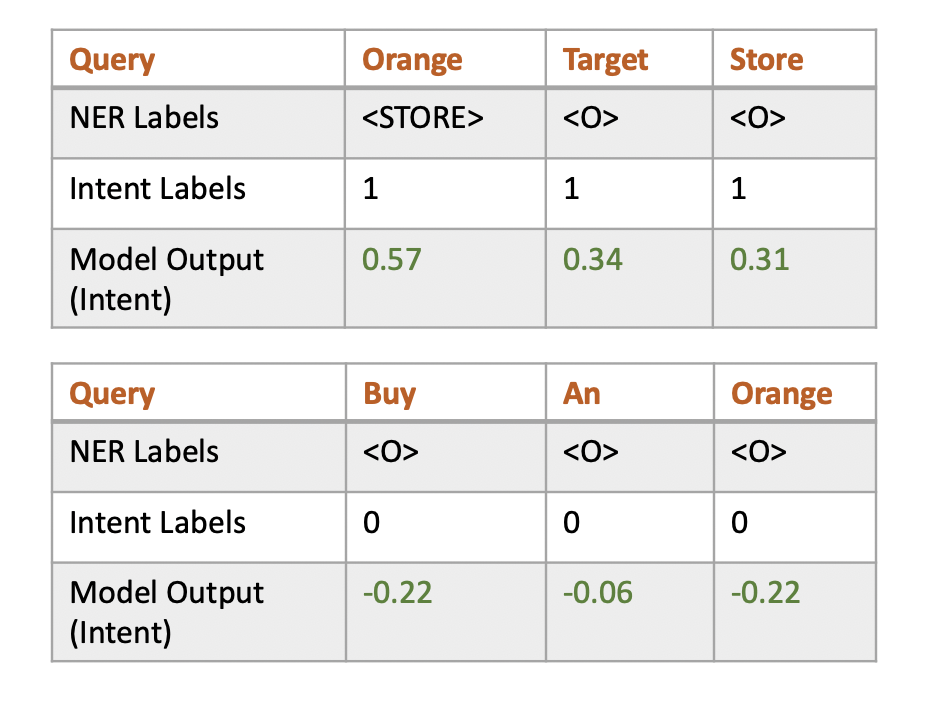}
\caption{NER and Intent Example}
\end{figure}
One loss function is the Softmax loss which classifies each word of a query into an entity type (store location name). This is a commonly used loss function for multi-label classification with known derivatives. The Softmax function is preceded by linear nodes and the Softmax is given as under:\\\\
$Softmax_i = \frac{exp(x_i)}{\sum_{j=1}^n exp(x_j)}$\\\\
The other loss function uses shared labels. If the number of words in a query is $n$, and if the intent of the query is store lookup, there are $n$ ones in the target labels. If the intent is not a store lookup, there are $n$ zeros in the target labels. The reason for having $n$ identical labels is that during inference, each word would have its own intent (which appears similar to the ideas in \cite{vaswani2017attention}). But the intent is shared by all the words of a query and it helps learning and improves the precision as well as the F-1 score, as table 1 shows. We compare our algorithm BERT-Multitask with BERT-base for the store lookup NER task.\\\\
So, there are two distinct forward and backward passes for the two cost functions. The passes are done separately and sequentially for the two losses. The learning rate for the global intent loss is reduced by a factor of $0.1$.\\\\
As we noted, the intent loss is shared during the backward pass, but during the forward pass, each word of the query has its own intent score. In our opinion, this facilitates learning and as the results show, it improves the precision and F-1 score.
\begin{figure}[ht]
\label{ner-algorithm}
\centering
\includegraphics[scale=0.4]{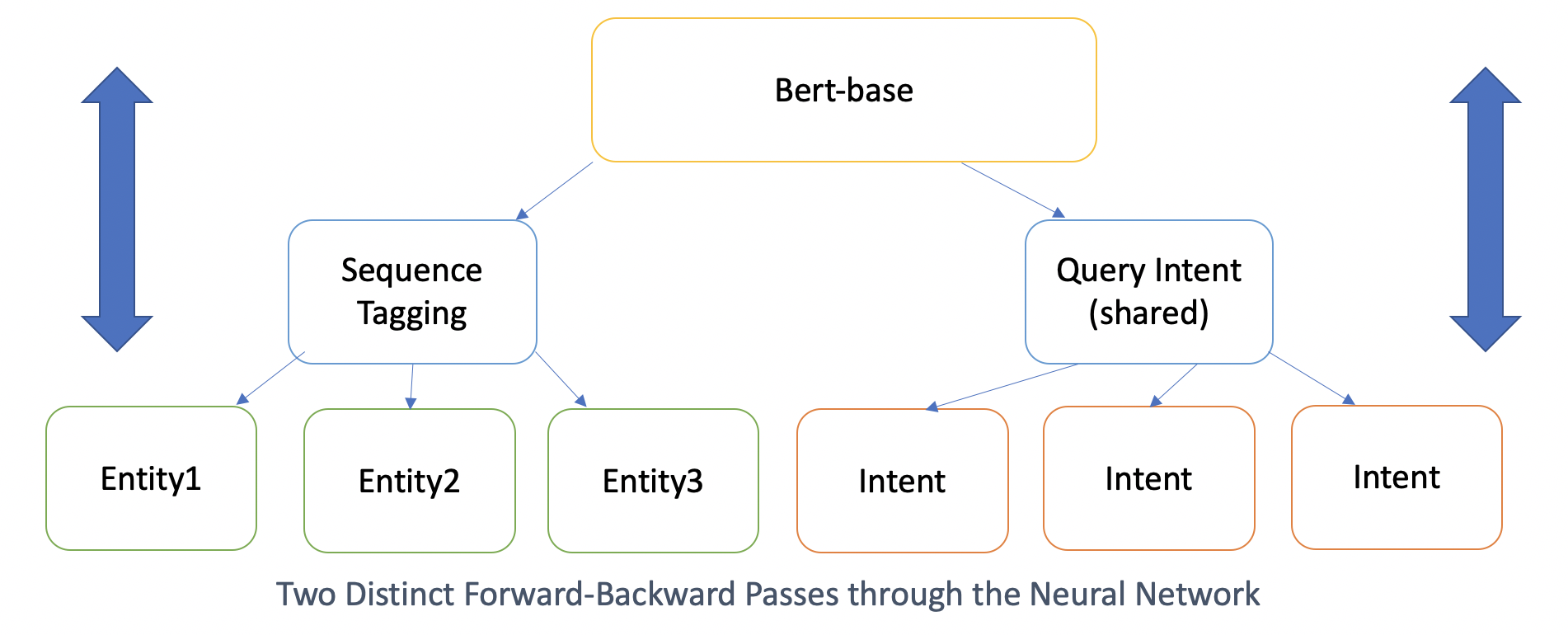}
\caption{Joint Learning of Intent and NER}
\end{figure}
\section{Results}
The results of our algorithm as compared to BERT-base are shown in table 1. Our algorithm is able to improve precision by $3\%$ by just doing an \textbf{argmax}. The goal of our algorithm is indeed, to improve precision and reduce false positives. At Target corporation, we deal with millions of distinct search/voice queries and a precision increase of $3\%$ is considered very significant and desirable.\\\\
Moreover, in our opinion, with algorithms improving the SOTA of NER constantly being developed, and with the SOTA reaching the $90's$, a $3\%$ increase in precision is difficult to achieve.\\\\
As seen in figure 1, the intent scores for the first query ("orange target store") are high for all three words. And for the second query ("buy an orange"), the intent scores are all negative. It is worth noting that the score for "orange" is significantly different between the two queries. So, what ends up happening is that if the intent is "store lookup", the scores for the entire query will be higher, but the interactions between intent and NER will push the scores for entities even higher. Working together, the intent and NER losses improve the results significantly.\\\\
We run both BERT-base and BERT-Multitask with the same random seed, to avoid variations because of the random number generators.
\begin{table}[t]
\caption{Results on a 20\% held out test set (same random seed)}
\label{results}
\begin{center}
\begin{tabular}{p{5cm}p{2cm}p{2cm}p{2cm}}
\hline
\bf Algorithm &\bf Precision &\bf Recall &\bf F-1 Score\\
\hline
BERT-Multitask &\bf 94\% &79\% &\bf 0.86\\
\hline
BERT-base &91\% &80\% &0.85\\
\end{tabular}
\end{center}
\end{table}
\section{Conclusion}
In this paper, we presented a multi-tasking BERT based named entity recognition algorithm which does task specific NER. Task specific NER, in our domain, means identifying store lookup intent with the store name. By using two loss functions, and different forward and backward passes, we find that the precision is improved.\\\\
Moreover, by having a separate intent score for each word in the query, thresholds can be computed (on a validation set) and classification can be done based on these threshold. But we haven't explored this yet.\\\\
Our algorithm is applicable to any task in which text classification precedes tagging. For instance, learning the intent of gifting and then tagging the gift in the query is another important application. There are numerous such applications.\\\\
Future work could include using the generated thresholds and see if the precision improves further. Another opportunity is to experiment with hyper-parameters (like the weight of the intent loss).
\nocite{*}
\bibliographystyle{unsrt}
\bibliography{ner}
\end{document}